\documentclass[letterpaper, 10 pt, conference]{ieeeconf}  

\IEEEoverridecommandlockouts                              

\usepackage{subfig}

\usepackage{graphicx} 
\usepackage{amsmath} 

\title{\LARGE \bf
DeepQ Stepper: A framework for reactive dynamic walking on uneven terrain\\}

\author{Avadesh Meduri$^{1}$, Majid Khadiv$^{2}$ and Ludovic Righetti$^{1,2}$
\thanks{This work was supported by New York University and the National Science Foundation (grant 1825993).}
\thanks{$^{1}$Tandon School of Engineering, New York University (NYU), USA. 
        {\tt\small am9789@nyu.edu, ludovic.righetti@nyu.edu}}%
\thanks{$^{2}$Max-Planck Institute for Intelligent Systems, Tuebingen, Germany.}%
}
\usepackage{subfig}
\usepackage{hyperref}

\begin{document}

\maketitle
\thispagestyle{empty}
\pagestyle{empty}

\begin{abstract}
Reactive stepping and push recovery for biped robots is often restricted to flat terrains because of the difficulty in computing capture regions for nonlinear dynamic models. In this paper, we address this limitation by using reinforcement learning to approximately learn the 3D capture region for such systems. We propose a novel 3D reactive stepper, The DeepQ stepper, that computes optimal step locations for walking at different velocities using the 3D capture regions approximated by the action-value function. We demonstrate the ability of the approach to learn stepping with a simplified 3D pendulum model and a full robot dynamics. Further, the stepper achieves a higher performance when it learns approximate capture regions while taking into account the entire dynamics of the robot that are often ignored in existing reactive steppers based on simplified models. The DeepQ stepper can handle non convex terrain with obstacles, walk on restricted surfaces like stepping stones and recover from external disturbances for a constant computational cost. 
\end{abstract}

\section{INTRODUCTION}
Development of fast contact and motion planning algorithms has been a topic of interest in robotics for many years. Fast contact planning algorithms will enable a robot to quickly make or break a contact with its environment in order to move to desired locations, recover from an external disturbance or react to a change in its surroundings. On the other hand, fast motion planning algorithms will enable a robot to compute in real-time a feasible trajectory from its current state or configuration to reach a desired contact configuration. Hence, in order to enable robots to generate reactive behaviors, we need to be able to adjust contact sequences and locations quickly along with adapting motion trajectories accordingly.

Most reactive motion planning algorithms, i.e. used for real-time trajectory generation, are based on simplified models such as the Linear Inverted Pendulum Model (LIPM) \cite{kajita20013d}. The LIPM along with predefined footstep locations was used to make a humanoid robot walk in \cite{wieber2006trajectory, kajita2003biped}. Since footstep locations are computed beforehand, these approaches cannot handle external disturbances that require the robot to adapt its footsteps to stabilize itself. These approaches have been extended to include contact location adjustment \cite{herdt2010online} but are limited to convex stepping regions and cannot easily handle obstacles. While nonlinear extensions have been proposed to handle non-convex stepping regions due to obstacles \cite{naveau2016reactive}, they still do not consider the full robot dynamics.

An important concept for foot step planning, the capture point \cite{pratt2006capture}, is the location where the robot should step in order to bring itself to rest after one step. The notion of capture point was further extended in \cite{koolen2012capturability} to capturable regions, which are regions where the robot is allowed to step so that there exists a sequence of steps after which the robot does not fall. An equivalent concept called the divergent component of motion (DCM), was proposed in parallel in \cite{takenaka2009real}. The DCM, also known as the instantaneous capture point (ICP), has been used to generate walking motions on planar \cite{takenaka2009real} and non planar ground with limited capability \cite{englsberger2013three}. 

In \cite{khadiv2020walking}, the notion of DCM offset is introduced to generate walking motions with desired velocities while being able to adapt both step location and timing online. Due to the linearity of the LIPM, it is possible to efficiently compute the next footstep location in real-time. However, this model assumes that the center of mass (CoM) dynamics is constrained to a plane and that the robot has no angular momentum. These assumptions do not hold anymore when walking on uneven terrains or when stepping fast enough such that the angular momentum effects are not negligible.

Contact planning algorithms that can handle complex terrains either utilize computationally expensive search algorithms, nonlinear optimization or mixed integer methods to generate feasible contact plans. Contact planners have long been limited to finding contact sequences that are statically feasible but have been recently extended to include dynamic feasibility constraints, increasing the range of possible movements. In \cite{mordatch2012discovery, posa2014direct}, the full-body motion and contact selection problems are formulated as single nonlinear optimization problems. While they can in principle find complex contact sequences, these approaches are computationally too
expensive to be used in realtime. In \cite{winkler2018gait}, a phase-based formulation of contact planning is proposed where the duration and location of each end-effector phase is optimized such that number of contact transitions along with the sum of the duration of all the contact phases is a constant. \cite{deits2014footstep} use mixed integer programming to formulate the contact planning program where integer variables are used to select end end-effector that makes a contact at a given time and to select a surface on which the end-effector makes contact. The optimizer only considers kinematic constraints to generate a feasible contact plan and the computation time grows exponentially with the number of terrains and steps in the plan. \cite{ponton2020efficient} introduce centroidal dynamics into the mixed-integer programming based contact planner so that dynamically feasible contact sequences and trajectories can be generated. \cite{lin2019efficient} generates dynamically feasible contact sequences using a variation of the $A^*$ algorithm with a learned contact feasibility classifier and heuristic prediction based on a centroidal dynamics planner \cite{ponton2020efficient}. Dynamic feasibility constraints are also considered in \cite{fernbach2020c} along with learned approximations of the centroidal dynamics to reduce computation time of the algorithm. Despite the reduced computation time of these contact planning algorithms, they are not quick enough to be used online to reactively generate contact plans for the robot. While all these approaches offer a certain level of generality, it remains challenging to use them for real-time control due to their computational complexity.

Given the LIPM assumption, to be able to walk without falling, it is necessary and sufficient to step within the (infinite-step) capturable region \cite{khadiv2020walking, koolen2012capturability}. Unfortunately, extending this result to more complicated scenarios including the robot nonlinear dynamics, uneven terrains or non-convex stepping regions is generally not feasible because of the intractability in computing the capture region in such cases. 

In this paper, we propose a novel reactive stepping approach, the DeepQ stepper, that approximates 3D capture regions for the full robot dynamics using deep reinforcement learning. We formulate the problem of choosing step locations as a continuous-state discrete-action Markov Decision Problem, affording the use of efficient Deep Q-Learning algorithms \cite{mnih2013playing}. The learned state-action value function (Q-function) implicitly encodes an approximation of the capture region which can then be used to plan footsteps online, at a constant computational cost, irrespective of the complexity of the terrain. Empirical evidence suggests that the DeepQ stepper learns a good approximation of both 2D  \& 3D capture regions even for nonlinear dynamic models. Extensive walking simulations with a biped robot with point feet demonstrate the advantages of our approach, which takes into account the full robot dynamics, including swing foot dynamics, acceleration limits, or joint friction. 

\section{Background}
In this section, we provide necessary background on the nonlinear inverted pendulum model,
results from capturability analysis and the DQN algorithm \cite{mnih2013playing}.
\subsection{nonlinear Inverted Pendulum} \label{sec:invertedpendulum}
The nonlinear inverted pendulum dynamics can be derived from the Newton-Euler equations of motion of a robot 
\begin{equation}
    m(\ddot{c} - g) = \sum_{i = 1}^{n} F_i
\end{equation}
\begin{equation}
    \dot{L} = \sum_{i = 1}^{n} (u_i - c) \times F_i
\end{equation}
where $m$ is the robot mass, $c$ is its CoM location, $g$ is the gravity vector, $F_i$ is the ground reaction force applied on leg $i$, $L$ is the angular momentum at the CoM and $u_i$ is the foot location (assuming a point contact foot). 
If we assume that there is no angular momentum around the CoM and the robot only makes contact with the ground with one foot at a given instant of time, the above equations simplify to the nonlinear pendulum dynamics
\begin{equation}
    c^{x, y} - \frac{c^{z}}{\ddot c^{z} + g^{z}}(\ddot c^{x,y} - g^{x,y}) = u^{x, y} \label{eq:invertedpendulumxy}
\end{equation}
\begin{equation}
    m(\ddot c^{z}+g^{z}) = F^{z} \label{eq:invertedpendulumz}
\end{equation}
The linear inverted pendulum dynamics \cite{kajita20013d} is obtained when $\ddot c^{z} = 0$. This renders the above equations linear and independent in $x$ and $y$. The $z$ direction is the vertical direction aligned with gravity.

\subsection{Capturability and Viability Regions}
For a walking robot, the N-step capturable region corresponds to the set of footsteps that would enable the robot to come to a stop after at most N steps. Computing the capture region for a general robot model is generally infeasible. However, the infinite step capture region for the linear inverted pendulum dynamics can be computed analytically \cite{koolen2012capturability}.
The DCM offset is defined as the distance between the next foot step and the DCM location at the time of next step. By ensuring that the DCM offset at each step remains in a certain bound \cite{khadiv2020walking}, it is possible to ensure that the robot remains capturable while also tracking a desired COM velocity.

\subsection{Reinforcement Learning}
One very successful deep reinforcement learning algorithm for continuous state space and discrete actions Markov decision processes (MDPs) is the Deep Q network (DQN) algorithm \cite{mnih2013playing}. The DQN algorithm uses a deep neural network (DNN) to approximate the Q-function and one DNN called the target network which closely follows the Q-function DNN \cite{lillicrap2015continuous}. The Q-function is then used to compute the optimal action for a given state. The algorithm aims to optimize the following least squares loss to train the function approximator
\begin{equation}
    \min_{\theta} \sum_{t} \left( Q_\theta(x_{t}, a_{t}) - c(x_t, a_t) - \alpha \min_{a} Q_{\theta^-}(x_{t+1}, a)\right)^2
\end{equation}
where $Q_\theta(x_{t}, a_t)$ is the DNN approximating the Q-function with weights $\theta$, $c(x_t, u_t)$ is the cost, $x_{t}$ is the state, $a_{t}$ is the action at time t, $\alpha \in (0,1)$ is a discount factor, $Q_\theta^-(x, a)$ is the target network DNN with weights $\theta^-$. 
In our implementation, we smoothly update the target network weights at each iteration $\theta^- = \tau \theta + (1-\tau) \theta^-$ as in \cite{lillicrap2015continuous}, where $\tau \in (0,1)$. Note that in this work, we minimize a cost function as opposed to maximizing a reward function.

\section{Approach: the DeepQ Stepper}
In this section we describe our approach to learn an efficient footstep planner.
Then, we provide details on our specific choices for the trajectory generation procedure and robot controller. Note however that our DeepQ Stepper does not rely on a specific underlying controller and it can be used with any other, more advanced, trajectory generation and control approaches.

\subsection{DeepQ Stepper} \label{sec:dqs}
The main idea of the DeepQ Stepper is to learn the value of each possible footstep given the current state of the robot, i.e. learn a Q-function associated to a stepping action, where the Q value represents the capability of the robot to either bring itself to rest or track a desired velocity after taking that step. The set of steps with low values will provide an approximation of the capture region. As we use a model-free approach to learn the Q-function, it can directly be used to approximate the capture region of the full robot dynamics. 

\subsubsection{Learning the value of a step}
During training, we will use either the (nonlinear) inverted pendulum dynamics (IPM) or the full robot dynamics as the model of the robot.
We discretize the possible stepping locations, such that we can use
the DQN approach. At the beginning of each step, the model is allowed to take an action, which corresponds to selecting one of the possible step lengths.
The state of the agent at a given time step consists of the distance of the CoM from the current foot location in each direction, the velocity of the CoM in the $x$ and $y$ directions, an index determining which foot is on the ground ($+1$ if right foot is on the ground, $-1$ if left foot is on the ground) and the desired velocity in the $x$ and $y$ directions. That is $x = [c_{x} - u_{x}, c_{y} - u_{y}, c_{z} - u_{z}, \dot{c}_{x}, \dot{c}_{y}, n, v_{x}^{des}, v_{y}^{des}]$.

The goal is to step so as to track a desired walking velocity, while trying to be as close as possible to the hip at the beginning of the next step and choosing as small step lengths as possible without falling down. The episode terminates after $n$ steps or if the robot falls down before the n steps. The agent is considered to have fallen down if the kinematic constraint (maximum allowed leg length) is violated. We use the following cost function
\begin{align}
    c(x,a) = w_{1}(|h^{x} - u^{x}| + |h^{y} - u^{y}|) \\
    + w_{2}(|\dot{c}_{x} - v_{x}^{des}| + |\dot{c}_{x} - v_{x}^{des}|) \nonumber\\
    + w_{3}(|a_{x}| + |a_{y}| + |a_{z}|) \nonumber
            + I(x)
\end{align}
where $w_{1}, w_{2}, w_{3}$ are weights, $c_{x,y,z}$ is the location of the CoM at the start of the next step, $h_{x,y,z}$ is the hip location of the leg whose foot is on the ground. $u_{x,y, z} = u^{0}_{x,y,z} + a_{x,y,z}$ is the foot location at the start of the next step, $u_{x,y,z}^{0}$ is the foot location at the start of the current step, $v_{x,y,z}^{des}$ is the desired velocity, $a_{x,y,z}$ is the chosen action at the start of the step and $I(x)$ is an indicator function that returns a constant value (here $100$) if the kinematic constraint is violated otherwise it returns zero. 

The DeepQ stepper architecture is designed so as to accept the state of the agent at the starting of each step along with an action (step location in the 3 directions) and return the corresponding Q-value. The optimal policy for the agent is to choose the action (step length) with the lowest Q-value. 

The DeepQ stepper is trained using the DQN algorithm described in the previous section with $w1=0.5$, $w2=3.0$, $w3=1.5$. To accelerate learning, we bias the exploration at the beginning of learning. We use an LIPM-based stepper to select actions 80 \% of the time to first fill the replay buffer. After the desired buffer size is reached, we use an $\epsilon-$greedy strategy with $\epsilon=0.2$.
 When sampling the mini-batch, we take 20 \% of the mini-batch to be the latest state-action pairs and the rest is sampled uniformly from the replay buffer. In our experiments, this significantly improved the rate of convergence of the algorithm especially because the rewards are sparse for the problem.  
 
 \subsubsection{The DeepQ Stepper as a reactive foot-step planner}
 During execution, the state of the agent is computed at the beginning of each step and provided to the trained DeepQ stepper.
 We compute the action with the lowest Q-value (i.e. the optimal step length) only using admissible steps given the current allowable stepping regions. Consequently, a constant number of network evaluations are needed for the stepper independently of the stepping region, which provides a significant computational advantage. The agent then takes the optimal action at the end of the step. This process is repeated indefinitely to generate walking motions on different terrain.

\subsection{Trajectory generation and control approach} 
We present the trajectory generation and control approach used to generate robot motions given a choice of footstep location. The approach is used during both learning and evaluation. Note however that the DeepQ stepper can be used with any other approach, as long as reasonable motions can be generated. It would in fact probably benefit from more advanced trajectory optimization methods for locomotion \cite{ponton2018time} at the cost of higher computational requirements.
\subsubsection{CoM trajectory generation}\label{sec:mp_walking}
We generate CoM trajectories using the nonlinear inverted pendulum dynamics, Eqs. \eqref{eq:invertedpendulumxy} and \eqref{eq:invertedpendulumz}. Initially, a trajectory optimization problem is solved for the dynamics governing the $z$ direction independently, such that the height of the inverted pendulum moves from its current height to the desired height by the end of the trajectory (duration of one step) while also reaching a zero velocity \cite{audren2014model,mirjalili2018whole}. The QP problem solved for the $z$ direction trajectory is
\begin{align} \label{eq:mp_formulation}
    \min. &\quad \sum_{i = 0}^{T}wF_{t}^{2} \\
    s.t. \quad &c^{z}_{t+1} = c^{z}_{t} + \delta t \dot{c}^{z}_{t} \nonumber\\
    &\dot{c}^{z}_{t+1} = \dot{c}^{z}_{t} + \delta t (F_{t}/m - g) \nonumber\\
    &c^{z}_{0} = h_{0}, \quad\ \dot{c}^{z}_{0} = 0 \nonumber\\
    &c^{z}_{T} = h_{T}, \quad \dot{c}^{z}_{T} = 0 \nonumber
\end{align}
where $c^{z}_{t}$ is the height at the time step t. $h_{0}, h_{T}$ are initial and final height of the inverted pendulum and $T$ is the number of collocation points in the trajectory optimization problem. The $z$ trajectory obtained from the motion planner is then plugged into inverted pendulum Eq. \eqref{eq:invertedpendulumxy} to integrate the $x$ and $y$ dynamics of the system to obtain a dynamically feasible CoM trajectory.

This method of trajectory planning is adopted to bypass the nonlinearity in the dynamics and subsequently generate feasible trajectories using a QP, which can be done very efficiently. This makes it possible to replan CoM trajectories while walking on uneven ground in the presence of external forces on the robot in real-time. Further, in the case where the height of the inverted pendulum is kept a constant, the motion plans from the solver coincide with the linear inverted pendulum, which will facilitate our experimental comparisons later. Note however that any other trajectory
generation approach could be used with the DeepQ stepper.

\subsubsection{Swing Foot Trajectory Generation} \label{sec:end_eff}
Feasible swing foot trajectories are generated for the robot at the beginning of each step after obtaining the desired next step location from the DeepQ stepper. The end-effector trajectory starts from the current location of the foot and ends at the next foot location. A 5th degree polynomial is used to parameterize the foot trajectory and the optimal parameters are obtained using quadratic programming. The QP solved at the beginning of each step is 
\begin{align}
    \min. \quad &\sum_{i = 0}^{5} k_{i}a_{i}^{2} \\
    s.t. \quad &f(0) = u_{0} \quad, \quad f'(0) = 0 \nonumber\\
    &f(T) = u_{next} \quad , \quad f'(T) = 0 \nonumber\\
    &f(T/2) = u_{via-point}\nonumber
\end{align}
where $f(t)$ is the $5^{th}$ order polynomial with parameters $a_{i}$ $(i = 0,...5)$ and $u_{0}, u_{next}$ are the current and next foot step location, $u_{via-point}$ is the desired mid-point of the step and $T$ is the step time. The solution to this small program can be computed very quickly. 

\subsubsection{Whole Body Controller} \label{sec:ctrl} 
The trajectory generated by the motion planner (subsection \ref{sec:mp_walking}) and the swing foot trajectory generator (\ref{sec:end_eff}) is tracked using the whole-body controller proposed in \cite{grimminger2020open}. It consists of two components. The first component tracks the desired center of mass trajectory by computing errors in the centroidal space and adding it to the feedforward forces computed by the motion planner. The total desired force in the centroidal space is then mapped to joint torques using the Jacobian between the hip and the foot on the ground. Subsequently, these torques are added to the joint torques computed by the second component of the controller that tracks a desired swing foot trajectory using a task space impedance controller. 

\section{Experiments}
We now present extensive simulation results to investigate the capabilities of the approach. In a first set of experiments, we evaluate the DeepQ stepper to control the simplest dynamic model, a 1D LIPM. This enables its analysis in light of exact results for the capturability region \cite{koolen2012capturability}.
Then, we evaluate the approach to generate walking for a 3D nonlinear inverted pendulum model. Finally, we demonstrate the ability of the approach to learn footstep planning using the full dynamics of a biped robot with point feet.
In particular, we show that the stepper is capable of learning the value of stepping regions that more adequately reflects the true capabilities of the robot. This leads to a better approximation of the capture region of the robot for each state and to an improvement of walking performance when compared to typical LIPM-based approaches.

\subsection{Experimental Setup}

For the simulations, we use Bolt (Fig. \ref{fig:scenarios}), a newly designed open source biped robot \cite{Daneshmand2020}, as we eventually aim to implement our approach on the real robot. Bolt is a torque-controlled biped robot with 6 active DoFs \cite{grimminger2020open, Daneshmand2020} capable of very dynamic walking. It has passive ankles which we model with point feet. Each leg is 0.4m long, its base is 0.13m wide and 0.078m high. It easily can step every 0.2 seconds \cite{Daneshmand2020}. All the modeling choices such as step time or nominal CoM height described below are based on the model of the real robot and its capabilities.

 The DeepQ stepper is trained using the procedure discussed in section \ref{sec:dqs}. The DNN of the DeepQ stepper contains 7 layers each containing 512 neurons followed by a final layer with one neuron. All layers except the last one are activated with a ReLU function. The stepper is trained with a learning rate of $1e-4$, $\tau=0.001$ and a buffer size of 8000. Each episode during training is terminated after 10 steps or when the agent falls down. The DNN architecture and training parameters are kept the same throughout all the experiments. All the training instances converged consistently for all the experiments we conducted.
 
 For all the experiments, we used a Dell precision 5820 tower machine with a 3.7 GhZ Intel Xeon processor. The DeepQ stepper was trained using PyTorch \cite{paszke2019pytorch} and we used PyBullet \cite{coumans2013bullet} for the robot simulation. The accompanying video\footnote{The accompanying video can also be found \href{https://www.youtube.com/watch?v=aITooZlm-WY}{here}} illustrate the simulation results.

\subsection{1D LIPM}
A 1D LIPM model is used to initially test the performance of the DeepQ stepper. The model has its CoM at a constant height of 0.35 m above the ground and is allowed to take a step every 0.2 seconds. The state of the agent is $x = [c_{x} - u_{x}, \dot{c_{x}}]$. We discretize the action space into 11 equally spaced step lengths starting from -0.4 m to 0.4 m (i.e. a 8 cm discretization step).

\begin{figure}
\vspace{0.3cm}
    \centering
    \includegraphics[scale = 0.2]{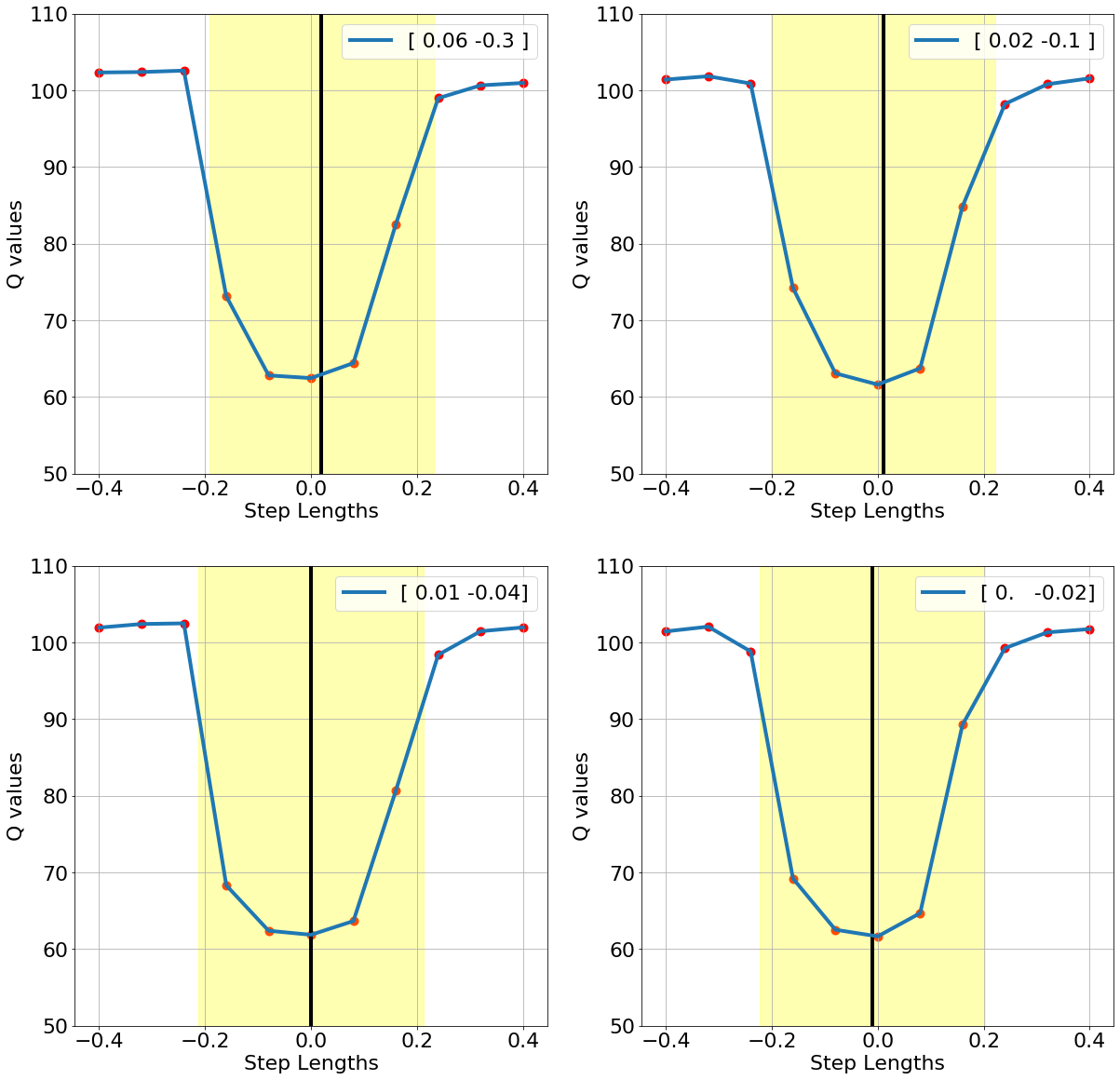}
   \caption{Plot of the Q-values for each action along with the location of the ICP/DCM (vertical black line) and the infinite step capture region (yellow region) for a 1D LIPM. Each plot shows a different state (blue label)}
    \label{fig:qvalues_vs_icp}
\end{figure}

The whole training process of the DeepQ stepper takes about 15 to 20 minutes (approximately 5000 episodes). A visualization of the Q values for different states of the agent at the start of the step is shown in Fig. \ref{fig:qvalues_vs_icp}. The location of the instantaneous capture point (ICP) at the end of the corresponding step along with the infinite step capture region is also shown. We notice that the optimal stepping strategy found by our approach is close to the location of the ICP for different states, which is the optimal stepping strategy to stop, for the LIPM model. Further, the value assigned to each action by the learned Q-function is such that the ones with the lower values lie within the capture region while those with higher values lie outside the capture region. This suggests that the Q-function has implicitly learned the capture region. During execution, the value of each step can be evaluated to select the next best possible step to take, so as to continue walking, in case the optimal action cannot be chosen (i.e. by choosing the step with the lowest value that is also feasible in the current environment). 

\subsection{3D Inverted Pendulum}
We now use our approach with a nonlinear inverted pendulum model (IPM) (Sec. \ref{sec:invertedpendulum}), for which no analytic capture region characterization exists. It contains a 0.13m-width base (the hips are 0.065m away from the CoM on the y axis), to learn to walk on uneven ground. The maximum leg length is 0.4m (same as the real robot), which introduces an additional kinematic constraint in the model and terminates the episode when violated. 
The model is expected to keep its CoM at a nominal height of 0.35 m above the ground at the start of each step and it follows the trajectory generated by the inverted pendulum motion planner (Sec. \ref{sec:mp_walking}). The model takes a step every 0.2 seconds and its state is the same as described in Sec. \ref{sec:dqs}. The action space now consists of 99 step lengths with 11 equally spaced step lengths in the forward direction ($x$ axis) from -0.4 m to 0.4 m and 9 geometrically spaced (geometric progression) step lengths in the lateral direction (y axis) starting from 0 m to 0.43 m. Since the minimum step length allowed in the lateral direction is zero, the model is not allowed to step to the right of the right foot and to the left of the left foot. This reduces the possibility of foot collision. However, the DeepQ stepper can be trained with an action space which allows for crossing of the feet, since the stepper can handle non convex reachable spaces efficiently. 

The DeepQ stepper takes approximately 6000 episodes to learn to walk without falling while tracking a desired velocity ranging from $\pm 0.7 m/s$. After learning, the stepper can recover balance without falling when an episode is initialised with velocities ranging from $\pm 1.0 m/s $.

\begin{figure}
\vspace{0.3cm}
    \centering
    \includegraphics[scale = 0.42]{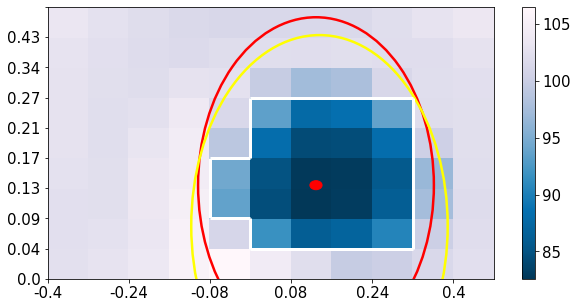}
    \caption{A visualization of the action values for the state [0, 0.065, 0.35, 0.16, -0.16, 1, 0, 0]. The infinite step capture region for the model at the end of the step is depicted by the red circle. The red dot shows the location of the ICP/DCM at the end of the step. The yellow circle represents the kinematic constraints at the end of the step.}
    \label{fig:qvalues_capture_region}
\end{figure}

A visualization of the Q values for a state = [0, 0.065, 0.35, 0.16, -0.16, 1, 0, 0] (The agent has a velocity of 0.16 m/s and -0.16 m/s in the x and y direction respectively, while the CoM is at a height of 0.35 m above the ground and 0.065 m to the left of the right foot) at the start of the step, on flat ground, along with the theoretical infinite step capture region, ICP/DCM and kinematic constraints for the model at the end of the step are shown in Fig. \ref{fig:qvalues_capture_region}. Stepping on flat ground is shown here, because the dynamics matches the LIPM, which enables comparison with theoretical results. Darker squares correspond to actions with lower cost (the darkest square is the optimal action according to the Q-function). The white box encloses the steps from which the robot can recover from. That is, if the IPM is forced to take any step within the white box (which need not be the best action) and after which it is allowed to take optimal actions, the DeepQ stepper is able to bring its CoM to rest without falling, i.e. is captured. This shows that the DeepQ stepper not only learns an optimal location to step for a given state (that is close to the ICP/DCM), but it is also able to learn an approximation of the true capture region for the nonlinear inverted pendulum dynamics. 

\begin{figure}
 \vspace{0.3cm}
    \centering
    \includegraphics[scale = 0.3]{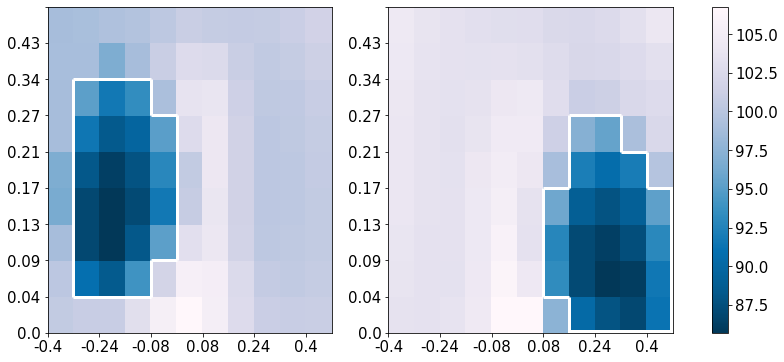}
    \caption{Heatmaps representing the action value of the learned 3D capture region for two state. The left figure is for the state x = [0, 0, 0.35, -0.5, 0.2, 1, 0, 0] and the agent is to step on a terrain that is 0.05 m below the current location of the foot. The right figure is for the state x = [0, 0, 0.35, 0.4, 0.0, 1, 0, 0] and all the next steps locations are on a terrain that is above the current COP location by 0.08 m}
    \label{fig:3d_q}
\end{figure}

During training, the IPM is also exposed to scenarios where the terrain varies randomly with heights ranging from $\pm 0.07 m$ (20 \% of the leg height of the robot while walking), so that the DeepQ stepper approximately learns the 3D capture region, which is the region in 3D space the IPM can step so as to remain capturable. Using the learned 3D capture region, the robot can navigate complicated terrain by accounting for terrain height while choosing optimal actions. Figure \ref{fig:3d_q} shows a visualization of the learned 3D capture region for two different scenarios. Since, the infinite step capture region for a nonlinear inverted pendulum cannot be computed analytically, it is not possible to compare the learned capture regions with the ground truth. However, we verified that the robot can walk on uneven terrain when steps within the white box of the heatmaps are taken, as in the case of 2D stepping discussed previously. It shows that the Q-function can be used to choose among a set of possible steps, at no additional cost, therefore approximating a 3D capture region and enabling foot step selection on complex terrains.

\subsection{3D walking with Bolt Robot}
Now we train the DeepQ stepper directly with a simulation of the complete robot to demonstrate that our algorithm can take into account its full dynamics such as inertia of the base, friction in the joints, swing foot dynamics, possible tracking errors in the controller, etc. All of which are ignored in the nonlinear inverted pendulum model.

\begin{figure*}
\vspace{0.3cm}
\centering
\includegraphics[width=17cm, height=3.6cm]{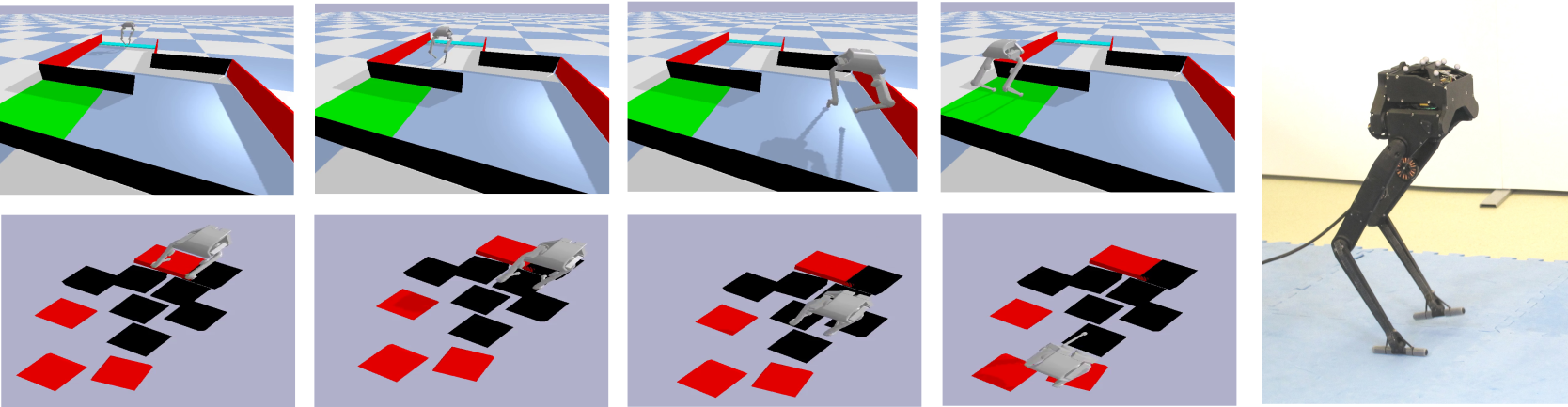}
\caption{Bolt navigating challenging environments with obstacles and stepping stones. Sequence of motion ordered starting from left to right.}
\label{fig:scenarios}
\vspace{-0.5cm}
\end{figure*}

\subsubsection{Comparison between the IMP and full-robot DeepQ stepper}
We compare the stepper learned with the 3D IPM with the
stepper learned in the full robot simulation to control
the full robot. Our goal is to show the improvement in footstep selection when taking into account the complete robot dynamics.
The 3D IPM stepper was not able to generate robust walking gaits with the real robot dynamics, due to significant effects of the swing leg. 
A similar issue appeared when a LIPM-based reactive planner that ignores the swing foot dynamics was transferred to the real robot \cite{Daneshmand2020}.
We therefore set a simulation where we reduced the mass of the legs to 10\% of the mass of the real robot.

Bolt is controlled using the procedure discussed in section \ref{sec:mp_walking}, \ref{sec:end_eff} and \ref{sec:ctrl}. Since the feet of Bolt are point contacts, it is not possible for the whole body controller to regulate the center of pressure(CoP) in the $x$ and $y$ direction beyond a point. Consequently, we do not control the CoM in horizontal directions and rely on optimal stepping to prevent the CoM from diverging. 

The DeepQ stepper trained on the 3D IPM is able to generate walking gaits while tracking velocities from $\pm 0.7$ m/s on flat terrain. It is able to recover from disturbances up to $\pm 0.8$ m/s, that is when the episode is initialised with these velocities, the stepper is able bring the robot to rest or track a desired velocity without falling down.  

The DeepQ stepper trained directly in the simulation environment with the modified Bolt robot (lighter legs) is trained in about 45 minutes (approximately 6000 episodes). The robot is controlled with the same control framework described above. It is also capable of generating robust walking gaits and recover from disturbances. 

\begin{figure}
    \centering
    \includegraphics[scale = 0.29]{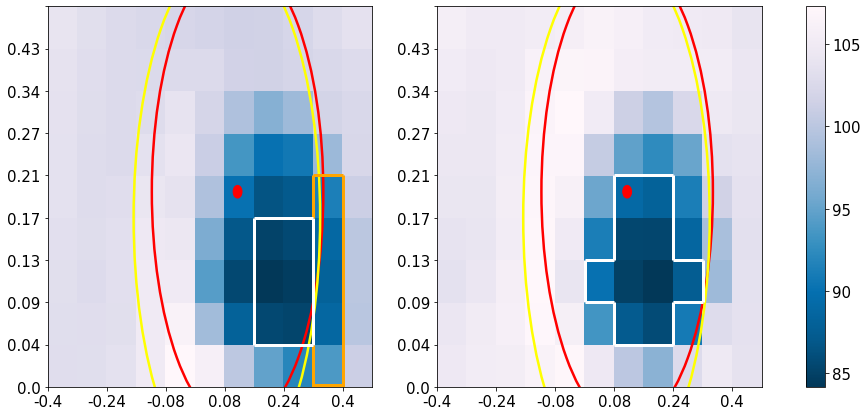}tha
    \caption{Comparison of 2D capture regions for DeepQ stepper trained in simulation (figure to the right) and in the inverted pendulum environment (figure to the left). The yellow circle represents the kinematic constraints, the red circle shows the infinite step capture region and the red dot shows the DCM/ICP.}
    \label{fig:bullet_vs_ipm}
\end{figure}

A comparison between the learned capture regions of the DeepQ stepper trained directly in simulation (right figure) and with the inverted pendulum environment (left figure) for the same state is shown in Fig. \ref{fig:bullet_vs_ipm}. The white box in the two heat maps encloses the step locations from which Bolt is able to recover by taking optimal actions after, when it starts from the same initial condition and uses the two DeepQ steppers in simulation. As can be seen, the DeepQ stepper trained in simulation is able to learn a larger capture region as compared to the inverted pendulum stepper. In addition, the orange box encloses actions that the IPM stepper believes are good actions to take, but in reality these actions lie outside the kinematic limits and can not be taken by the robot. This discrepancy arises because the rate of divergence of the CoM of the robot is different from the simplified IPM for the same state. Subsequently, actions that are viable with the IPM for the same state, are not feasible on the robot. In contrast, the simulation stepper returns high Q values for actions that lie outside the kinematic region, since it learns to step using the full robot dynamics.

\begin{figure}
    \centering
    \includegraphics[scale = 0.38]{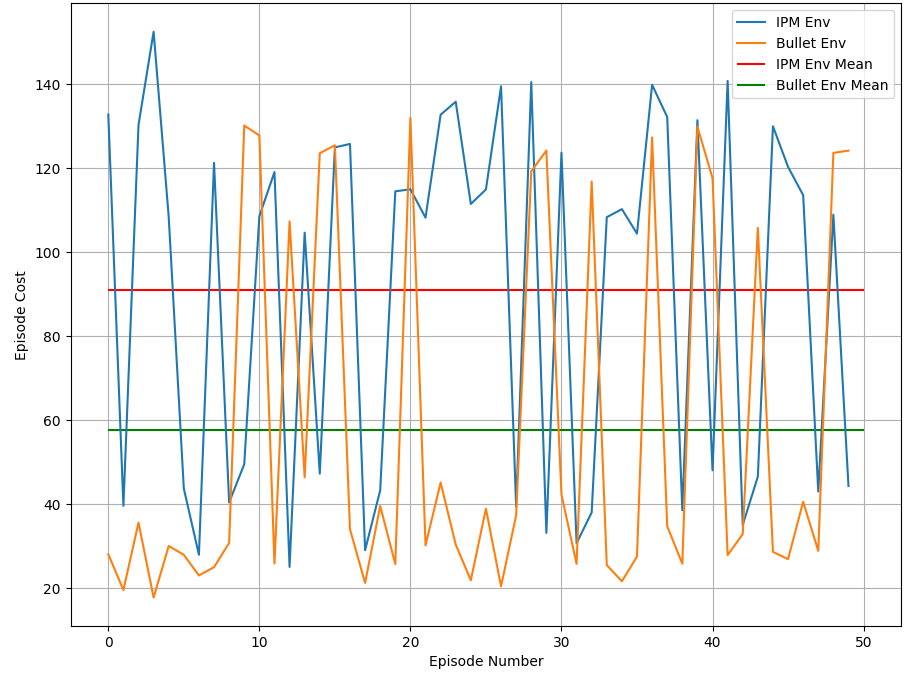}
    \caption{Comparison of episode cost during roll out in simulation environment with DeepQ stepper trained in the inverted pendulum env (IPM Env) and directly in simulation (Bullet Env).}
    \label{fig:episode_comparison}
\end{figure}

Figure \ref{fig:episode_comparison} shows a comparison of episode costs over 50 episodes in the simulation environment. During each episode, the robot is initialised with random conditions (different CoM states, desired tracking velocities and uneven terrain) and both controllers are executed for 15 steps. The stepper trained in the simulation environment shows a better performance about 77\% of the time (lower episode cost) when tested over 1000 episodes. The IPM stepper often fails to capture the robot in scenarios with high CoM velocities where the swing foot dynamics becomes prominent because this requires taking large steps to recover. Consequently, the stepper does not account for this and chooses steps that are longer than necessary which destabilises the robot. On the other hand, the simulation DeepQ stepper is able to account for these dynamics and learn a richer representation of the capture region (the Q values are quite different within the white box, as one moves away from the optimal action), where it prefers to take smaller steps whenever possible to improve the stability of the robot and reduce the affects of the swing foot dynamics and still recover from high initial CoM velocities. In addition, the simulation DeepQ stepper is more robust while walking on uneven terrain and is able to track desired velocities better as compared to the IPM stepper. This can also be seen in Fig \ref{fig:episode_comparison} where the bullet stepper incurs lower episode cost in episodes where both the steppers are able to generate walking without falling down (episodes with cost under 100).

\subsubsection{DeepQ stepper with robot dynamics}
 The DeepQ stepper is now trained with the real robot dynamics. During training, the simulation environment is initialised with a sparsely generated terrain consisting of random step heights up to 0.05 m (contains both flat and non convex terrain). After every 500 episodes of training the terrain is made more complicated by adding more non convex steps into the environment. The DeepQ stepper takes about 6000 episodes to converge.
 
\begin{figure}
\vspace{0.3cm}
    \centering
    \includegraphics[scale = 0.3]{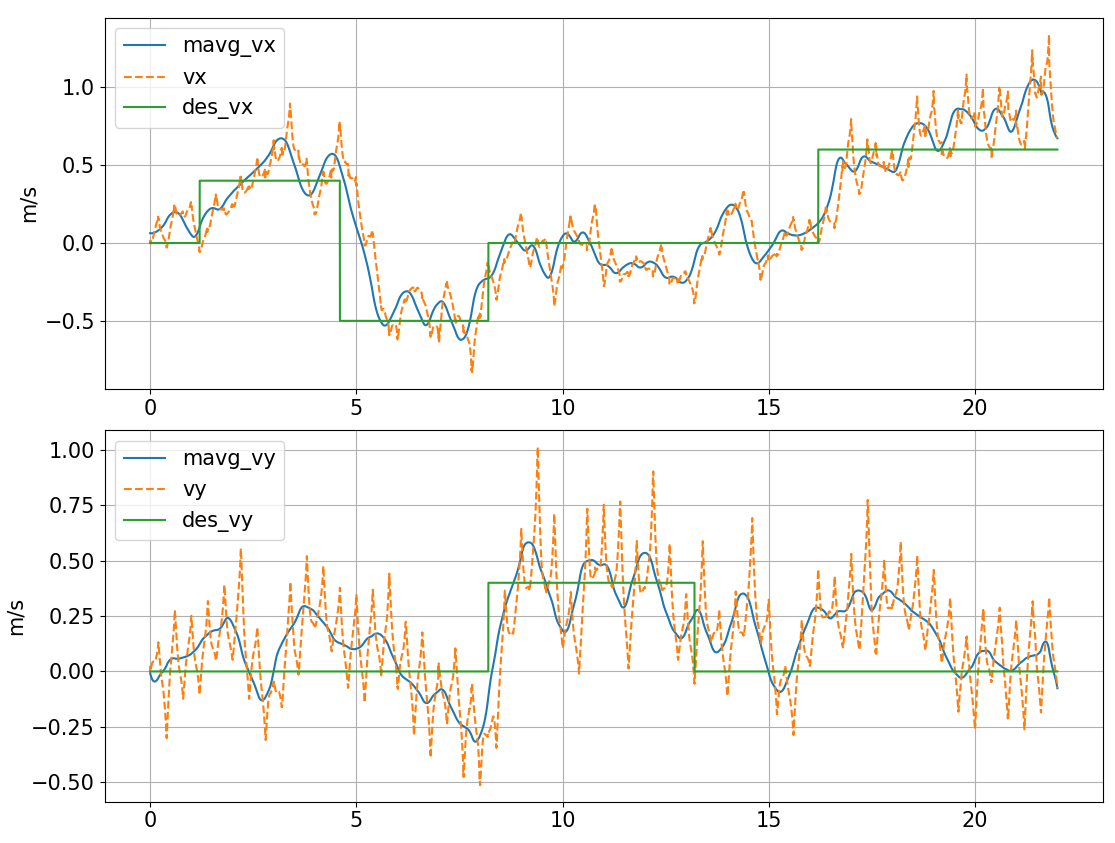}
    \caption{CoM velocity, COM moving average of 0.2 second and desired velocity plot that shows the tracking capability of the DeepQ stepper}
    \label{fig:3D_obstacle_avoidance}
\end{figure}

The DeepQ stepper trained directly in the simulation environment is able to navigate complex terrains (Fig. \ref{fig:scenarios}). The robot is able to track and change desired velocities online to avoid obstacles reasonably well, despite having point feet and discrete allowed step lengths (Fig. \ref{fig:3D_obstacle_avoidance}). The DeepQ stepper is also able to navigate uneven terrains while the robot tracks desired velocities in different directions. Since the stepper is not affected by the density or nature of the terrain (convex or non convex) it is able to provide optimal step locations in real time. In addition, since the terrain is not flat, the CoM height is constantly changing while the robot walks. LIPM approximations that are used in existing reactive steppers would not hold in such scenarios and it becomes important to account for the nonlinear dynamics of the system, which the DeepQ stepper is able to do.

In situations where the robot can only step in certain regions such as stepping stones, the DeepQ stepper is able to provide information to the robot about the next best actions to take when it is not possible to choose the optimal action because it learns a capture region. Subsequently, it becomes possible to plan foot step locations on complex terrain online, with no additional computational cost. Finally, the stepper is able recover from external disturbance ranging from $\pm 3 N$ (push recovery) on both flat and non convex, uneven terrain. The performance of the DeepQ stepper while walking on complex terrain, stepping stones and recovering from external disturbances is shown in the attached video.

\section*{Conclusion}
In this work, a novel reactive stepping framework, the DeepQ stepper is proposed. The stepper can learn the 3D capture regions for nonlinear systems such as the nonlinear inverted pendulum and a full biped robot for which analytical solutions do not exist. The stepper uses this information to decide optimal locations to step on uneven terrain while tracking a desired CoM velocity. The low computation time of the stepper makes it suitable to react to external disturbances. The DeepQ stepper is able to provide alternative step options in scenarios where optimal actions can not be taken as in the case of stepping stones. Further, the stepper is not affected by non-convex terrain and its computation time remains the same irrespective of the number of steps in the environment. When trained with the robot in simulation, the DeepQ stepper is able to learn a non trivial capture region that improves its performance as compared to learning the capture region using the inverted pendulum agent which ignores swing foot dynamics, joint frictions etc.

In future work, we aim to train the DeepQ stepper directly on the real robot to learn walking by accounting for the entire dynamics of the robot. Further, we seek to extend the approach to more complicated motions such as running gaits using centroidal dynamics planners.

\bibliography{Master}
\bibliographystyle{IEEEtran}

\end{document}